\documentclass[sigconf]{acmart}
\usepackage{hyperref}
\usepackage{colortbl}
\usepackage{graphicx}
\usepackage{float}

\usepackage{latexsym}
\usepackage{url}
\usepackage{xspace}
\usepackage{graphicx}
\usepackage{multirow}
\usepackage{makecell}
\usepackage{lscape}
\usepackage{bm}
\usepackage{algorithm}
\usepackage{algpseudocode}
\usepackage{framed}
\usepackage{color}
\usepackage{xcolor}
\usepackage{booktabs,colortbl}
\usepackage{bbm}
\usepackage{subfigure}
\usepackage{multirow}

\usepackage{amssymb}
\usepackage{amsmath}
\usepackage{multirow}
\usepackage{makecell}
\usepackage{tabularx}
\usepackage{booktabs}
\usepackage{setspace}
\usepackage{utfsym}
\usepackage{pifont}
\usepackage{arydshln}
\definecolor{lightgrayv}{HTML}{EEEDF3}
\definecolor{redv}{HTML}{C00000}
\definecolor{bluev}{HTML}{0070C0}

\newcommand{\eg}{\emph{e.g.,}\xspace}
\newcommand{\ie}{\emph{i.e.,}\xspace}
\newcommand{\etc}{\emph{etc.}\xspace}

\newcommand{\baby}{\textsc{Misder}\xspace}
\AtBeginDocument{%
  }

\setcopyright{acmlicensed}
\copyrightyear{2025}
\acmYear{2025}
\acmDOI{XXXXXXX.XXXXXXX}

\acmConference[CIKM '25]{The 34th ACM International Conference on Information and Knowledge Management}{10 - 14 November, 2025}{Seoul, Korea}
\acmISBN{978-1-4503-XXXX-X/18/06}

\begin{document}

\title{Variety Is the Spice of Life: Detecting Misinformation with Dynamic Environmental Representations}

\author{Bing Wang}
\orcid{0000-0002-1304-3718}
\affiliation{
  \institution{College of Computer Science and Technology, Jilin University}
  \city{Changchun}
  \state{Jilin}
  \country{China}}
\email{wangbing1416@gmail.com}

\author{Ximing Li}
\authornote{Ximing Li and Bo Yang are the corresponding authors. All authors are also affiliated with Key Laboratory of Symbolic Computation and Knowledge Engineering of the Ministry of Education, Jilin University.}
\orcid{0000-0001-8190-5087}
\affiliation{
  \institution{College of Computer Science and Technology, Jilin University}
  \city{Changchun}
  \state{Jilin}
  \country{China}}
\email{liximing86@gmail.com}

\author{Yiming Wang}
\orcid{0000-0001-6867-670X}
\affiliation{
  \institution{College of Computer Science and Technology, Jilin University}
  \city{Changchun}
  \state{Jilin}
  \country{China}}
\email{yimingw17@gmail.com}

\author{Changchun Li}
\orcid{0000-0002-8001-2655}
\affiliation{
  \institution{College of Computer Science and Technology, Jilin University}
  \city{Changchun}
  \state{Jilin}
  \country{China}}
\email{changchunli93@gmail.com}

\author{Jiaxu Cui}
\orcid{0000-0002-4922-0915}
\affiliation{
  \institution{College of Computer Science and Technology, Jilin University}
  \city{Changchun}
  \state{Jilin}
  \country{China}}
\email{cjx@jlu.edu.cn}

\author{Renchu Guan}
\orcid{0000-0002-7162-7826}
\affiliation{
  \institution{College of Computer Science and Technology, Jilin University}
  \city{Changchun}
  \state{Jilin}
  \country{China}}
\email{guanrenchu@jlu.edu.cn}

\author{Bo Yang}
\authornotemark[1]
\orcid{0000-0003-1927-8419}
\affiliation{
  \institution{College of Computer Science and Technology, Jilin University}
  \city{Changchun}
  \state{Jilin}
  \country{China}}
\email{ybo@jlu.edu.cn}

\renewcommand{\shortauthors}{Wang et al.}

\begin{abstract}
  The proliferation of misinformation across diverse social media platforms has drawn significant attention from both academic and industrial communities due to its detrimental effects. Accordingly, automatically distinguishing misinformation, dubbed as \textbf{M}isinformation \textbf{D}etection (\textbf{MD}), has become an increasingly active research topic. The mainstream methods formulate MD as a static learning paradigm, which learns the mapping between the content, links, and propagation of news articles and the corresponding manual veracity labels. However, the static assumption is often violated, since in real-world scenarios, the veracity of news articles may vacillate within the dynamically evolving social environment. To tackle this problem, we propose a novel framework, namely \textbf{Mis}information \textbf{D}etection with \textbf{D}ynamic \textbf{E}nvironmental \textbf{R}epresentations (\textbf{\baby}). The basic idea of \baby lies in learning a social environmental representation for each time period and employing a temporal model to predict the representation for future periods. 
In this work, we specify the temporal model as the LSTM model, continuous dynamics equation, and pre-trained dynamics system, suggesting three variants of \baby, namely \baby-\textsc{lstm}, \baby-\textsc{ode}, and \baby-\textsc{pt}, respectively.
To evaluate the performance of \baby, we compare three variants to various MD baselines across two prevalent MD datasets, and the experimental results can indicate the effectiveness of our proposed model.

\end{abstract}

\ccsdesc[500]{Computing methodologies~Artificial intelligence}
\ccsdesc[500]{Information systems~Social networks}

\keywords{social media, misinformation detection, time series prediction}


\maketitle

\section{Introduction}

Social media platforms, \eg Twitter and Weibo, have nowadays emerged as the main medium for distributing news articles in real life. These platforms provide more convenient access to news, but they also bring the persistent issue of fake ones, posing risks to various domains, \eg societal credibility and economy \cite{difonzo2013rumor,wang2019systematic}.
For example, the dissemination of fake claims regarding the \textit{Russia-Ukraine} war has instilled panic within the global community, resulting in significant repercussions for their import and export trade.\footnote{\url{https://www.bbc.com/news/60513452}} 
Consequently, to prevent the widespread of misinformation, automatically distinguishing them, dubbed as \textbf{M}isinformation \textbf{D}etection (\textbf{MD}), becomes the capital challenge of the big picture of misinformation control, aiming to eliminate the negative effects of misinformation. Naturally, MD has recently gained considerable attention from both academic and industrial communities \cite{zhang2021mining,zhu2022generalizing,sheng2022zoom}.

\begin{table}[t]
\centering
\renewcommand\arraystretch{1.25}
\small
  \caption{Two examples illustrating the dynamic evolution of the veracity of news. They show that the veracity of the same news piece can vacillate in different social environments.}
  \label{example}
  \setlength{\tabcolsep}{5pt}{
  \begin{tabular}{m{8.0cm}}
    \toprule
    \ding{202} \textbf{News}: {\fontsize{9pt}{10pt}\selectfont Trump received the majority in the Electoral College and won upset victories in the Rust Belt region. The pivotal victory in this region, which Trump won by less than 80,000 votes in the three states, was considered the catalyst that won him the Electoral College vote.} \\
    \rowcolor{lightgrayv} \textbf{Veracity label}: {\color{bluev} Real (2016)} $\boldsymbol{\Rightarrow}$ {\color{redv} Fake (2020)} \\
    \hline
    \ding{203} \textbf{News}: {\fontsize{9pt}{10pt}\selectfont The government announced at a press conference that we are currently in a period of high incidence of respiratory infectious diseases, the peak period of influenza and the second wave of new pneumonia, which has led to an increase in the number of fever patients currently seeking treatment.} \\
    \rowcolor{lightgrayv} \textbf{Veracity label}: {\color{bluev} Real (2021)} $\boldsymbol{\Rightarrow}$ {\color{redv} Fake (2023)} \\
    \bottomrule
  \end{tabular} }
\end{table}

Over the past decade, numerous MD arts have been proposed and achieved remarkable performance \cite{nan2021mdfend,chen2022cross,hu2023learn}. Generally, these methods predominantly focus on detecting misinformation by leveraging the semantic features of news content in offline datasets \cite{ma2018rumor,ying2023bootstrapping,zhu2023memory}. To further enhance detection accuracy, they often incorporate a range of external features inspired by various psychological and sociological theories, \eg emotional signals \cite{zhang2021mining}, publisher intents \cite{wang2023understanding,wang2024why}, and entity knowledge \cite{dun2021kan,zhu2022generalizing}, from news to provide additional contextual cues to enhance the performance of MD.
Typically, these approaches implement detection by manually annotating a \textbf{static} veracity label, \eg fake and real, for each news article. This labeled data is then used to train deep learning models that learn the underlying patterns between the news content and its corresponding veracity label \cite{zhang2021mining,ying2023bootstrapping,wang2024why}. By identifying these patterns, the models aim to generalize effectively to new, unseen data.

Despite these static regimes achieving encouraging success, they are often overly idealist in real-world scenarios. This is mainly because the static assumption is often violated. The recent sociological literature presents, beyond the intrinsic characteristics of the news, the \textbf{social environment} also plays a role in affecting the veracity of news \cite{de2021s,gurgun2023we}. Referring to Cambridge Dictionary,\footnote{\url{https://dictionary.cambridge.org/}} the social environment is defined as:
``\textit{all the facts, opinions, etc. relating to a particular thing or event}.''  
However, the social environment always \textbf{dynamically} evolves over periods. Therefore, the veracity of any news article may vacillate within such a dynamic social environment. As two representative examples shown in Fig.~\ref{example}, the veracity can be changed as the evolution of facts and public opinions about the presidential election and the coronavirus pandemic.

Motivated by the aforementioned analysis, we aim to indirectly obtain dynamic veracity by learning the evolution of the social environment over time.
To achieve this goal, we propose a novel MD framework, named \textbf{Mis}information \textbf{D}etection with \textbf{D}ynamic \textbf{E}nvironmental \textbf{R}epresentations (\textbf{\baby}). The basic idea behind \baby is to learn a social environmental representation for each time period and employ a time series model to explicitly capture the evolving patterns of these representations, and then predict their future states.
Specifically, we take inspiration from the recent advancements in prompt tuning techniques \cite{li2021prefix,lester2021the} and incorporate a learnable prompt into a pre-trained misinformation detector. This prompt, referred to as \textbf{D}ynamic \textbf{E}nvironmental \textbf{R}epresentation (\textbf{DER}), serves as a representation of the current social environment.
To make DER dynamic, we split the original MD dataset into several subsets according to their time stamps. Based on these subsets, we can train a specific DER for each time period. These period-specific DERs are then treated as ground-truth labels to learn a time series prediction model that captures their temporal dynamics.
We further introduce three variants of our proposed method based on the used time series model, namely \textbf{\baby-\textsc{lstm}}, \textbf{\baby-\textsc{ode}}, and \textbf{\baby-\textsc{pt}}. In detail, the \baby-\textsc{lstm} variant employs a standard LSTM model for temporal prediction, capturing sequential dependencies between DERs. On the other hand, due to the evolution of the social environment being consistently continuous, we draw inspiration from \cite{zang2020neural,zhang2024up2me} to present \baby-\textsc{ode}, which learns a neural dynamics equation for continuous-time predictions. Additionally, we propose a pre-trained \baby-\textsc{pt} variant, leveraging pre-trained models for enhanced generalization in predicting evolving social environments.

We conduct an evaluation of our proposed method \baby using two time-specific benchmark datasets \textit{Weibo} \cite{sheng2022zoom} and \textit{GossipCop} \cite{shu2020fakenewsnet}. These datasets allow us to simulate a realistic temporal setting by training our model on past-period data and testing it on future-period data. Through comprehensive experiments, we demonstrate the superior performance of \baby variants compared to current state-of-the-art MD methods. Additionally, we conduct an in-depth exploration of the continuous prediction capabilities of the \baby-\textsc{ode} variant including long-term prediction analysis and a case study. \textit{Our source code and data are released in \url{https://github.com/wangbing1416/MISDER}}. 

Overall, our contributions can be summarized into the following three aspects:
\begin{itemize}
    \item In this study, we address the issue of the dynamic social environment and propose a novel MD framework \baby.
    \item Our framework aims to fully leverage social environmental representations by employing a prompt tuning technique. To capture the evolution of these representations, we introduce three variants: \baby-\textsc{lstm} based on the LSTM model, \baby-\textsc{ode} utilizing continuous-time dynamics equations, and \baby-\textsc{pt} using pre-trained dynamics systems.
    \item Extensive experimental results demonstrate the significant impact of our proposed framework \baby in improving the performance of MD models. Moreover, we thoroughly investigate the continuous nature exhibited by \baby-\textsc{ode} throughout the analysis.
\end{itemize}

\section{Related Work} \label{sec2}

In this section, we review and summarize some related literature on misinformation detection, time-series prediction, and neural dynamics equations.

\subsection{Misinformation Detection}

In general, the primary goal of a misinformation detector is to identify fake and misleading content disseminated on social media platforms. Over the past decade, most MD studies have focused on analyzing the inherent semantic content of news to distinguish between real and fake information. Specifically, some studies have developed MD methods designed to process news articles containing only textual information, and they enhance their models by incorporating, such as entity knowledge \cite{dun2021kan,zhu2022generalizing}, sentiment signals \cite{zhang2021mining,luvembe2023dual}, publisher intents \cite{wang2023understanding,wang2024why}, and event information \cite{wang2018eann,zhang2024evolving}, into their detectors. Many of these approaches draw inspiration from psychological and sociological theories, leveraging specialized tools to extract associated features and integrating them with the semantic representations of news content. For example, BERT-EMO \cite{zhang2021mining} highlights the importance of emotional cues conveyed by both the publishers of the news and the reactions of the crowd, therefore, it captures the dual emotional features to enhance MD.
In addition, with the emergence of Large Language Model (LLM) technology, some recent studies have utilized LLMs to enhance existing MD models through data augmentation, \eg generating knowledge-intensive textual rationales \cite{hu2024bad}, simulating user responses to generate user comments \cite{wan2024dell,nan2024let}, and improving retrieval-augmented generation systems to utilize external evidence \cite{wang2024explainable}.

In addition to textual information, social media platforms often host multimodal content, including text, images, and videos. Therefore, to address the detection of misinformation including text and image pairs, some recent works design multimodal MD models to capture and integrate features of different modalities \cite{wang2018eann,khattar2019mvae,chen2022cross,wang2023cross,wang2024escaping}. For example, \textsc{Hami-m}$^3$\textsc{d} \cite{wang2024harmfully} enhances multimodal MD by capturing image manipulation artifacts and intent-related features, which are then integrated with multimodal semantic features to improve detection accuracy.
Another line of MD works involves considering the propagation structure of news pieces on social media. These approaches treat a news article, along with its associated comments and reposts, as a tree structure, and employ a variety of graph-based models, \eg graph convolutional networks \cite{kipf2017semi,li2024graph}, to capture their hidden embeddings \cite{ma2018rumor,lin2023zero}.

In this study, we focus on the temporal evolution of misinformation that received limited attention in existing research. To our knowledge, only \citet{zhu2022generalizing} and \citet{hu2023learn} attempt to mitigate the entity and topic bias resulting from shifts in the training data collected from different time periods. However, their focus is solely on the evolution of the distributions in entities and topics, and they do not provide an explicit definition of temporal evolution as a time series. 
Moreover, recent MD works also involve the consideration of social context. However, it is important to note that their notion of social context primarily revolves around users, comments, and retweets of a tweet, rather than our broader understanding including social events, \etc Specifically, they incorporate social context to explore the evolutions of user stances towards specific news events to address the challenge of detecting emerging misinformation. To achieve this, these works introduce a variety of time series methods using LSTM models \cite{li2022adadebunk,sun2022ddgcn,zuo2022contunually}.

\subsection{Time Series Prediction and Dynamics}

Previous time series prediction works treat the time series data as a sequence of vectors and use various deep models, \eg CNN and RNN, to capture their temporal dependency \cite{lai2018modeling,wu2020connecting}. Recently, with the potential of Transformer models \cite{vaswani2017attention} in sequence prediction tasks, some studies have begun improving their architecture to reduce computational complexity and expand the input window of the attention mechanism \cite{zhou2021informer,wu2021autoformer}. Meanwhile, some studies have started exploring the performance of Transformer-based pre-trained language models \cite{devlin2019bert,raffel2020exploring} on time-series data \cite{gruver2023large} and have proposed a variety of adaptation methods to generalize these language models to time-series tasks \cite{chang2023llm,jin2024time}.

Although time series prediction is a classic task with remarkable performance, its discrete nature makes it challenging to operate effectively in continuous states. This issue can be addressed by dynamics equations.
Typically, a dynamics equation is formalized as an Ordinary Differential Equation (ODE), and its goal is to model the continuous-time evolution of real-world systems, such as physical and genetic systems \cite{gao2016universal}. Recently, there has been increasing attention from the community towards neural dynamics equations with the advancement of deep learning techniques. These approaches aim to learn an ODE using data-driven methods \cite{chen2018neural,huang2023generalizing,wu2024prometheus}. For instance, NeuralODE \cite{chen2018neural} takes inspiration from residual networks \cite{he2016deep}, and enables an end-to-end training of an ODE system using a neural network and a black-box \texttt{ODESOLVER}. Another approach, NDCN \cite{zang2020neural}, implements NeuralODE on complex networks and proposes an encoding-decoding process.

\section{The Proposed Model} \label{sec3}

\begin{figure*}[ht]
  \centering
  \includegraphics[scale=1.22]{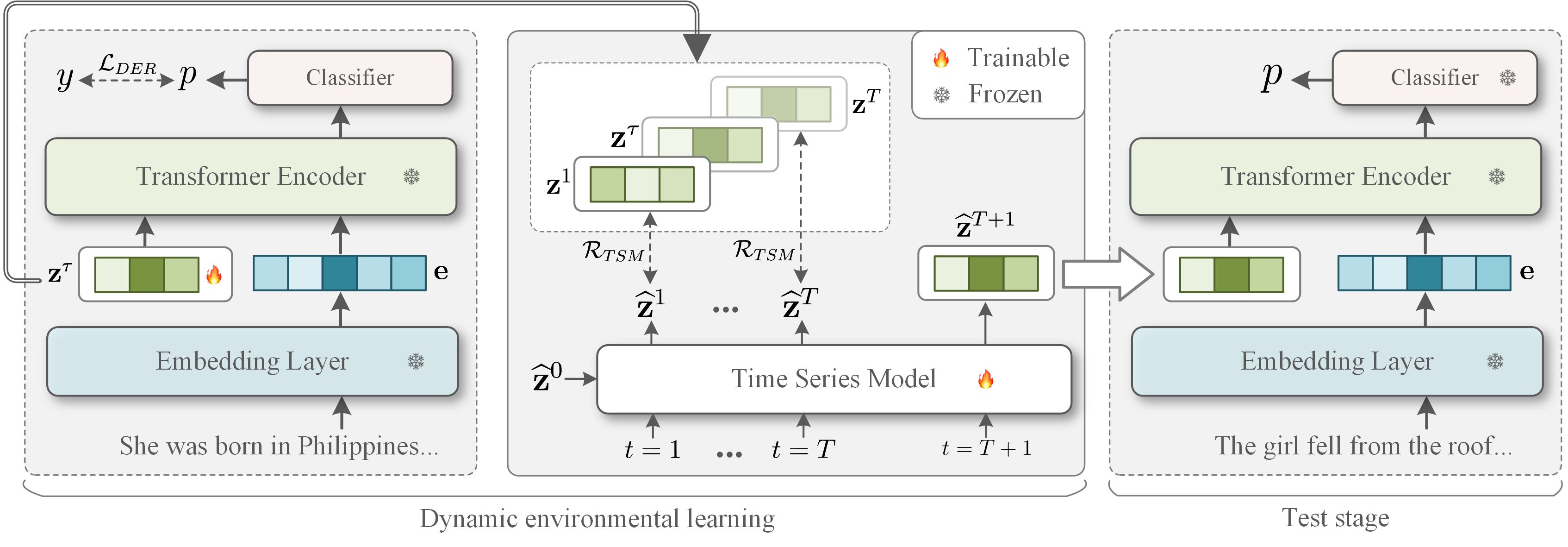}
  \caption{The overall framework of \baby. Our main target is to learn a DER for each period and predict them for future periods. Accordingly, we split the overall MD dataset into multiple subsets according to their time stamps, and use them to train DERs $\{\mathbf{z}^\tau\}_{\tau=1}^T$ for each subset. Based on these DERs, we learn a time series model specified by the LSTM model or neural dynamics equation to predict a new DER for the future.}
  \label{framework}
\end{figure*}

In this section, we introduce the brief task definition of time-evolving MD and our proposed \textbf{\baby} method in more detail.

\vspace{2pt} \noindent
\textbf{Task definition.}
In MD, the training dataset of $N$ samples is typically denoted as $\mathcal{D} = \{\mathbf{x}_i, y_i\}_{i=1}^N$, where $\mathbf{x}_i$ indicates the $i$-th news article and $y_i \in \{0, 1\}$ represents its corresponding veracity label, \ie \textit{fake} or \textit{real}. The primary target of MD is to induce a misinformation detector to give accurate veracity predictions for unseen news articles.
In our work, we concentrate on the time-evolving MD setting. To accomplish this, we extend the training dataset to $\mathcal{D} = \{\mathbf{x}_i, y_i, t_i\}_{i=1}^N$, where $t_i$ represents the time stamps of $\mathbf{x}_i$. Then, $\mathcal{D}$ is utilized to supervise the misinformation detector, enabling it to predict the veracities of unseen news articles whose time stamps are situated in the future of $\mathcal{D}$ \cite{zhu2022generalizing}. For clarity, the important nonations and their descriptions are listed in Table~\ref{notation}.

\begin{table}[t]
\centering
\renewcommand\arraystretch{1.10}
\small
  \caption{Summary of notations and their descriptions.}
  \label{notation}
  \setlength{\tabcolsep}{5pt}{
  \begin{tabular}{m{1.8cm}<{\centering}|m{5.8cm}<{\centering}}
    \bottomrule
    Notation & Description \\
    \hline
    $\mathbf{x}_i \in \mathcal{D}$ & $i$-th news article in the dataset \\
    $y_i \in \{0, 1\}$ & veracity label of $\mathbf{x}_i$ \\
    $t_i$ & time stamp of $\mathbf{x}_i$ \\
    $\{\mathbf{z}^\tau\}_{\tau=1}^T$ & DERs in $T$ time periods \\
    $N$ & number of samples in $\mathcal{D}$ \\
    $K$ & length of DER \\
    $T$ & number of temporal subsets of $\mathcal{D}$ \\
    $\mathcal{G}_{\boldsymbol{\Pi}}(\cdot)$ & pre-trained embedding layer \\
    $\mathcal{F}_{\boldsymbol{\theta}}(\cdot)$ & feature extractor \\
    $f_{\boldsymbol{\Phi}}(\cdot, t)$ & time series model \\
    $\mathbf{W}_C$ & veracity classifier \\
    \bottomrule
  \end{tabular} }
\end{table}

\subsection{Model Overview}
In this study, we present a novel \baby framework to address the issue of dynamic social environment. 
Generally, the training process of our \baby consists of two stages: \textbf{model warm-up} and \textbf{dynamic environment learning}. For clarity, the overall framework of \baby is demonstrated in Fig.~\ref{framework}.

The model warm-up stage learns an initial misinformation detector using the dataset $\mathcal{D}$. Specifically, given a piece of news article $\mathbf{x}_i = \{w_{ij}\}_{j=1}^L \in \mathcal{D}$, where $L$ indicates the news length, we utilize a pre-trained embedding layer $\mathcal{G}_{\boldsymbol{\Pi}}(\cdot)$, \eg the embedding matrix of BERT \cite{devlin2019bert}, to generate its word embedding $\mathbf{e}_i = \mathcal{G}_{\boldsymbol{\Pi}}(\mathbf{x}_i) \in \mathbb{R}^{L \times D}$. 
Our primary motivation is to capture the temporal evolution of the social environment of news. To achieve this, we introduce a \textbf{D}ynamic \textbf{E}nvironmental \textbf{R}epresentation (\textbf{DER}). Inspired by the recent prompt tuning technique \cite{li2021prefix}, one DER is initialized as a learnable matrix, denoted as $\mathbf{z}^0 \in \mathbb{R}^{K \times D}$, where $K$ represents the prompt length which is empirically fixed to 32. Then, $\mathbf{z}^0$ and $\mathbf{e}_i$ are concatenated and jointly fed into an arbitrary news feature extractor $\mathcal{F}_{\boldsymbol{\theta}}(\cdot)$ to get the veracity prediction $p_i$ of $\mathbf{x}_i$ as follows:
\begin{equation}
    \label{eq1}
    p_i = \mathcal{F}_{\boldsymbol{\theta}} \left( [\mathbf{z}^0 ; \mathbf{h}_i] \right) \mathbf{W}_C,
\end{equation}
where $[\cdot\ ; \cdot]$ indicates the concatenation operation, and $\mathbf{W}_C$ denotes a veracity classifier implemented by a typical feed-forward neural network, which contains two linear layers and two ReLU activation functions. Accordingly, the training objective of the model warm-up stage is as follows:
\begin{equation}
    \label{eq2}
    \mathop{\boldsymbol{\min}} \limits_{\boldsymbol{\Pi}, \boldsymbol{\theta}, \mathbf{W}_C, \mathbf{z}^0} \mathcal{L}_{MWU} = \frac{1}{N} \sum \nolimits _{i=1}^N \ell_{BCE} (p_i, y_i), 
\end{equation}
where $\ell_{BCE}(\cdot\ , \cdot)$ denotes the binary cross-entropy loss.

Afterward, the dynamic environment learning stage involves training different DERs for distinct periods, and uses them to learn a time series model for predicting a new DER for the future period. Specifically, we first split the original MD dataset $\mathcal{D}$ into its $T$ subsets, denoted as $\{\mathcal{D}^\tau\}_{\tau=1}^{T}$, based on their respective time stamps $t_i$. Given the subset $\mathcal{D}^\tau$ corresponding to the $\tau$-th period, we freeze the parameters of the embedding layer $\boldsymbol{\Pi}$, the misinformation detector $\boldsymbol{\theta}$ and the veracity classifier $\mathbf{W}_C$, and exclusively optimize the DER, \ie $\mathbf{z}^\tau$ with the objective as follows:
\begin{equation}
    \label{eq4}
    \begin{aligned}
        \mathop{\boldsymbol{\min}} \limits_{\{\mathbf{z}^\tau\}_{\tau=1}^T} \mathcal{L}_{DER}
    = \frac{1}{T} \sum \nolimits _{\tau=1}^T \frac{1}{|\mathcal{D}^\tau|} \sum \nolimits _{i=1}^{|\mathcal{D}^\tau|}  \ell_{BCE} (p_i^\tau, y_i^\tau), \\
        p_i^\tau = \mathcal{F}_{\boldsymbol{\theta}} \left( [\mathbf{z}^\tau\ ; \mathbf{e}_i^\tau] \right) \mathbf{W}_C, \quad 
    \mathbf{e}_i^\tau = \mathcal{G}_{\boldsymbol{\Pi}}(\mathbf{x}_i^\tau), \quad \mathbf{x}_i^\tau \in \mathcal{D}^\tau,
    \end{aligned}
\end{equation}
where $|\mathcal{D}^\tau|$ indicates the number of the samples in $\mathcal{D}^\tau$. To specify $\mathbf{z}^\tau$ to represent the social environment of the MD data in the $\tau$-th time period, we design a strategy to initialize it before being optimized by $\mathcal{L}_{DER}$. To achieve this, referring to the definition of the social environment ``\textit{the facts about a particular event},'' we first use an off-the-shelf large language model, \eg GPT-3 \cite{brown2020language}, to extract an event set $\mathcal{E}^\tau$ of the articles in $\mathcal{D}^\tau$, with manually designed prompts such as ``\textit{Extract the event from the article}''. Then, we input $\mathcal{E}^\tau$ into the embedding layer $\mathcal{G}_{\boldsymbol{\Pi}}(\cdot)$ to obtain their hidden representations and average these representations as the initialization of $\mathbf{z}^\tau$.

Given the optimized $\{\mathbf{z}^\tau\}_{\tau=1}^T$, we further consider them as the ground-truth labels, which are used to train a time series model $f_{\boldsymbol{\Phi}}(\cdot, t)$.
The objective of optimizing ${\boldsymbol{\Phi}}$ is represented as:
\begin{equation}
    \label{eq5}
    \mathop{\boldsymbol{\min}} \limits_{\boldsymbol{\Phi}} \mathcal{R}_{TSM},
\end{equation}
where $\mathcal{R}_{TSM}$ will be introduced in the following sections. During the dynamic environment learning stage, Eqs.~\eqref{eq4} and \eqref{eq5} are optimized alternatively.
Then, by utilizing this time series model, we can obtain the predicted DER $\mathbf{\widehat z}^{T + 1} = f_{\boldsymbol{\Phi}}(\mathbf{z}, T + 1)$ for the future testing time period $T + 1$.
Accordingly, we can develop a new misinformation detector to provide accurate predictions for the news articles posted during this future period. For clarity, we summarize the overall implementations of \baby in Alg.~\ref{alg1}. 

In the subsequent sections, we will describe two specific variations \baby-\textsc{lstm}, \baby-\textsc{ode}, and \baby-\textsc{pt} of our proposed method based on the structures of the time series models.

\subsection{Discrete-time \baby-\textsc{lstm}}
For this variant, we utilize a widely used discrete LSTM model to implement the time series prediction model.
Specifically, given a sequence of DERs $\{\mathbf{z}^0, \mathbf{z}^1, \cdots, \mathbf{z}^{\tau - 1}\}$, we employ an LSTM model \cite{sak2014long} as the time series prediction model to generate the prediction of current time step $\tau$:
\begin{equation}
    \label{eq6}
    \mathbf{\widehat z}^\tau = f_{\boldsymbol{\Phi}} \left( [\mathbf{z}^0, \mathbf{z}^1, \cdots, \mathbf{z}^{\tau - 1}] \right).
\end{equation}

Therefore, the training objective of the dynamic environment learning step in Eq.~\eqref{eq5} is formulated as follows:
\begin{equation}
    \label{eq7}
    \mathop{\boldsymbol{\min}} \limits_{\boldsymbol{\Phi}} \mathcal{R}_{TSM} = \frac{1}{T} \sum \nolimits _{\tau=1}^T \|\mathbf{\widehat z}^\tau - \mathbf{z}^\tau\|,
\end{equation}
where $\|\cdot\|$ is the L1 loss function.

\subsection{Continuous-time \baby-\textsc{ode}}

For this variant, a continuous-time neural dynamics equation is used to implement the time series model.
In contrast to the discrete-time \baby-\textsc{lstm}, the evolutions of the social environment always exhibit continuous and emergent patterns over time. Therefore, we introduce a continuous-time variant \baby-\textsc{ode} to capture the dynamics. Specifically, \baby-\textsc{ode} is designed to learn the following neural dynamics equation $f_{\boldsymbol{\Phi}}(\cdot, t)$, which represents the time-dependent differential of the DER.
\begin{equation}
    \label{eq8}
    \frac{\mathrm{d} \mathbf{z}}{\mathrm{d} t} = f_{\boldsymbol{\Phi}}(\mathbf{z}, t).
\end{equation}

Using this dynamics equation, we can calculate the predicted DER $\mathbf{\widehat z}^\tau$ at an arbitrary and continuous time $\tau$ by employing the following formulation:
\begin{equation}
    \label{eq9}
    \mathbf{\widehat z}^\tau = \mathbf{\widehat z}^0 + \int _0^\tau f_{\boldsymbol{\Phi}}(\mathbf{z}, t) \mathrm{d} t,
\end{equation}
where $\mathbf{\widehat z}^0$ represents the initial state of a dynamics function, and we fix $\mathbf{\widehat z}^0 = \mathbf{z}^0$.

To enable the random optimization of the parameter ${\boldsymbol{\Phi}}$, we draw inspiration from NDCN \cite{zang2020neural} to design our dynamics implementation. The specific optimization objective can be formulated as follows:
\begin{align}
    \label{eq10}
    \mathop{\boldsymbol{\min}} \limits_{\boldsymbol{\Phi}} \quad & \mathcal{R}_{TSM} = \frac{1}{T} \int _0^T \|\mathbf{\widehat z}^\tau - \mathbf{z}^\tau\| {\mathrm{d} t}, \\
    \textbf{s.t.} \quad & \mathbf{z}_h^0 = f_{\mathbf{W}_E}(\mathbf{\widehat z}^0), \nonumber \\
    & \mathbf{z}_h^\tau = \mathbf{z}_h^0 + \int _0^\tau f_{\boldsymbol{\Phi}}(\mathbf{z}_h, t) \mathrm{d} t, \nonumber \\
    & \mathbf{\widehat z}^\tau = f_{\mathbf{W}_D}(\mathbf{z}_h^\tau), \nonumber
\end{align}
where $f_{\mathbf{W}_E}(\cdot)$ and $f_{\mathbf{W}_D}(\cdot)$ are two feed-forward neural networks to encode $\mathbf{\widehat z}^0$ into the hidden space and decode the predicted version $\mathbf{\widehat z}^\tau$ back, respectively.
We calculate the dynamics using the \texttt{ODESOLVER} implemented with the widely-adopted \textit{dopri5} strategy \cite{wanner1996solving}:
\begin{equation}
    \label{eq11}
    \int _0^\tau f_{\boldsymbol{\Phi}}(\mathbf{z}_h, t) \mathrm{d} t = \texttt{ODESOLVER}(\mathbf{z}_h, \tau, {\boldsymbol{\Phi}}).
\end{equation}

To simplify the integral in Eq.~\eqref{eq10}, we use the pre-defined subsets $\{\mathcal{D}^\tau\}_{\tau=1}^{T}$ and formulate Eq.~\eqref{eq10} as Eq.~\eqref{eq7}.

\renewcommand{\algorithmicrequire}{\textbf{Input:}}
\renewcommand{\algorithmicensure}{\textbf{Output:}}
\begin{algorithm}[t]
    \caption{Training summary of \baby.}
    \label{alg1}
    \begin{algorithmic}[1]
    \Require Training dataset $\mathcal{D}$; iteration numbers $E_w$ and $E_d$ of two steps.
    \Ensure Parameters of the misinformation detector $\Omega = \{\boldsymbol{\Pi}, \boldsymbol{\theta}, \{\mathbf{z}^\tau\}_{\tau=1}^T, \mathbf{W}_C\}$ and the time series model ${\boldsymbol{\Phi}}$.
    \State Initialize $\boldsymbol{\Pi}$ and $\boldsymbol{\theta}$ with the pre-trained BERT model, and randomly initialize the other parameters.
    \State {\color{gray} // Step 1: model warm-up}
    \For{$i = 1, 2, \cdots, E_w$}
        \State Draw a mini-batch $\mathcal{B}$ from $\mathcal{D}$ randomly.
        \State Optimize $\Omega$ with $\mathcal{L}_{MWU}$ in Eq.~\eqref{eq2}.
    \EndFor
    \State {\color{gray} // Step 2: dynamic environment learning}
    \For{$i = 1, 2, \cdots, E_d$}
        \State Draw a subset $\mathcal{D}^\tau$ from $\mathcal{D}$.
        \State Draw a mini-batch $\mathcal{B}^\tau$ from $\mathcal{D}^\tau$ randomly.
        \State Calculate $\mathbf{\widehat z}^\tau$ with Eq.~\eqref{eq6} or Eq.~\eqref{eq9}.
        \State Alternatively optimize $\mathbf{z}^\tau$ with $\mathcal{L}_{DER}$, and ${\boldsymbol{\Phi}}$ with $\mathcal{R}_{TSM}$.
    \EndFor
    \end{algorithmic}
\end{algorithm}

\subsection{Pre-trained \baby-\textsc{pt}}

Typically, a dynamics system is solely effective on one specific dataset. Inspired by existing pre-training techniques \cite{chang2023llm,jin2024time}, we use a dynamics system pre-trained on multiple dynamics data \cite{chang2023llm}, which is intialized by a pre-trained language model \cite{raffel2020exploring}, and induce a new variant \baby-\textsc{pt}. This approach can not only capture the semantics present in DER, but also learn its temporal evolution.

Specifically, similar to Eq.~\eqref{eq10}, we replace the original FFNN layer $\mathbf{W}_E$ with a convolutional layer and a T5 encoder layer \cite{raffel2020exploring}, and replace $\mathbf{W}_D$ with a T5 decoder layer. Since T5 is trained on general-purpose text generation tasks, this dynamical model is inherently semantic-aware. Upon this, we follow \citet{chang2023llm} and further pre-train the dynamical model using dynamics data to enable it to capture temporal dependency awareness. Formally, the pre-training data is denoted as a time series $\mathbf{z}_p = \big[ \mathbf{z}_p^I;\mathbf{z}_p^O \big]$, where a sequence is split into an input state sequence $\mathbf{z}_p^I$ and an output state sequence $\mathbf{z}_p^O$. The pre-training objective is to reconstruct $\mathbf{z}_p^I$ and forecast $\mathbf{z}_p^O$ as follows:
\begin{align}
    \label{eq12} 
    \mathop{\boldsymbol{\min}} \limits_{\mathbf{W}_E, \mathbf{W}_D, \mathbf{W}_R^I, \mathbf{W}_R^O}
    \big\| & f_{\mathbf{W}_D} \big( f_{\mathbf{W}_E}(\mathbf{z}_p^I) \big) \mathbf{W}_R^I, \mathbf{z}_p^I \big\|  \nonumber \\
    & + \big\| f_{\mathbf{W}_D} \big( f_{\mathbf{W}_E}(\mathbf{z}_p^I) \big) \mathbf{W}_R^O, \mathbf{z}_p^O \big\|,
\end{align}
where $\mathbf{W}_R^I$ and $\mathbf{W}_R^O$ are different linear heads for reconstructing and forecasting tasks. Based on the pre-trained models $f_{\mathbf{W}_E}(\cdot)$ and $f_{\mathbf{W}_D}(\cdot)$, we follow Eq.~\eqref{eq10} to calculate the dynamics of DERs and their optimization objective.

\section{Experiments} \label{sec4}

In this section, We empirically evaluate and analyze the proposed \textbf{\baby} method specified by three variants.

\subsection{Experimental Settings}

\noindent
\textbf{Datasets.}
We conduct our experiments on two prevalent MD datasets \textit{Weibo} \cite{sheng2022zoom} and \textit{GossipCop} \cite{shu2020fakenewsnet}. 
\begin{itemize}
    \item The \textbf{\textit{Weibo}} dataset \cite{sheng2022zoom} is a collection of Chinese news articles spanning from 2010 to 2018. In this work, we focus on a challenging dynamic MD setting. Therefore, following previous works \cite{zhu2022generalizing}, we split the dataset into training, validation, and test sets. Specifically, the news articles posted between 2010 and 2017 are assigned to the training set, and the ones posted in 2018 are split as the validation and test sets. 
    \item For the \textbf{\textit{GossipCop}} dataset \cite{shu2020fakenewsnet}, we also split it according to the time stamps. Its training set includes articles published from 2000 to 2017, while the test set consists of articles published in 2018.
\end{itemize}

For clarity, the statistics of two datasets are shown in Table~\ref{dataset}.

\vspace{2pt} \noindent
\textbf{Baselines.}
We compare our \baby with dozens of MD baselines, and their brief descriptions are listed as follows:
\begin{itemize}
    \item \textbf{EANN} \cite{wang2018eann} is a classical multimodal MD model that employs an adversarial network to extract event-invariant features for addressing emergent MD. Following \cite{zhu2022generalizing}, we discard the image modality to adapt the setting in this paper.
    \item \textbf{MDFEND} \cite{nan2021mdfend} is dedicated to multi-domain adaptive MD and presents a mixture-of-expert model comprising multiple domain experts.
    \item \textbf{BERT-EMO} \cite{zhang2021mining} integrates emotional signals into the MD model by extracting sentiment features from text contents, thereby enhancing the detection of misinformation.
    \item \textbf{BERT} \cite{devlin2019bert} is a prevalent language model that is pre-trained on a large corpus, and we employ it as the backbone detector to evaluate the plug-and-play baselines and \baby.
    \item \textbf{ENDEF} \cite{zhu2022generalizing} stands out as the sole MD study that specifically focuses on the temporal evolution of misinformation articles. It is a plug-and-play method that can be combined with various backbone models, \eg BERT.
    \item \textbf{DER} \cite{li2021prefix} represents an MD model that incorporates a learnable DER $\mathbf{z}$ into the BERT model.
\end{itemize}

\begin{table}[t]
\centering
\renewcommand\arraystretch{1.05}
  \caption{Statistics of two prevalent MD datasets.}
  \label{dataset}
  \setlength{\tabcolsep}{5pt}{
  \begin{tabular}{m{2.05cm}<{\centering}m{0.7cm}<{\centering}m{0.7cm}<{\centering}m{0.7cm}<{\centering}m{0.7cm}<{\centering}m{0.7cm}<{\centering}m{0.7cm}<{\centering}}
    \toprule
    \multirow{2}{*}{Dataset} & \multicolumn{2}{c}{\# Training} & \multicolumn{2}{c}{\# Validation} & \multicolumn{2}{c}{\# Test} \\
    \cmidrule(r){2-3} \cmidrule(r){4-5} \cmidrule(r){6-7}
    & Fake & Real & Fake & Real & Fake & Real \\
    \hline
    \textit{Weibo} \cite{sheng2022zoom} & 2,561 & 7,660 & 499 & 1,918 & 754 & 2,957 \\
    \textit{GossipCop} \cite{shu2020fakenewsnet} & 2,024 & 5,039 & 604 & 1,774 & 601 & 1,758 \\
    \bottomrule
  \end{tabular} }
\end{table}

\vspace{2pt} \noindent
\textbf{Implementation details.}
For the backbone models BERT, we utilize the publicly available \textit{chinese-bert-wwm-ext} \cite{cui2020revisiting}\footnote{Downloaded from \url{https://huggingface.co/hfl/chinese-bert-wwm-ext}.}  for the Chinese dataset \textit{Weibo} and \textit{bert-base-uncased} \cite{devlin2019bert}\footnote{Downloaded from \url{https://huggingface.co/bert-base-uncased}.} for the English dataset \textit{GossipCop}. \baby-\textsc{pt} initializes a T5 model \textit{t5-base} \cite{raffel2020exploring}\footnote{Downloaded from \url{https://huggingface.co/google-t5/t5-base}.} without instruct tuning to keep its pre-trained temporal capability. To optimize these pre-trained BERT models while maintaining computational efficiency, we follow the previous approach \cite{zhu2022generalizing} by freezing the parameters of the first ten Transformer layers \cite{vaswani2017attention} and fine-tuning only the final layer. The sequence length $K$ of DER is empirically set to 32.
During the training stage, we implement an early stop mechanism to prevent overfitting. This mechanism halts training if the Macro F1 score on the validation set does not improve for 5 consecutive epochs, ensuring that the model generalizes effectively without unnecessary overtraining. We employ the Adam optimizer \cite{kingma2015adam} with a learning rate of $7 \times 10^{-5}$ for BERT and the veracity classifier, $1 \times 10^{-5}$ for DER and $1 \times 10^{-3}$ for the time series model, and the batch size is consistently fixed to 64. 
For the implementation of the \texttt{ODESOLVER}, which is integral to our \baby-\textsc{ode} and \baby-\textsc{pt} variants, we use the open-access repository provided by \citet{chen2018neural}.\footnote{Downloaded from \url{https://github.com/rtqichen/torchdiffeq}.}

\definecolor{grayv}{HTML}{707070}
\begin{table*}[t]
\centering
\renewcommand\arraystretch{1.02}
  \caption{Experimental results of our \baby on two prevalent datasets \textit{Weibo} and \textit{GossipCop}. All reported scores are reproduced by us. The bold and underlined fonts indicate the best and second best scores. The results marked by * indicate that the performance of \baby are statistically significant than the baseline BERT model (paired t-test $<$ 0.05).}
  \label{result}
  \setlength{\tabcolsep}{5pt}{
  \begin{tabular}{m{2.55cm}m{1.5cm}<{\centering}m{1.5cm}<{\centering}m{1.5cm}<{\centering}m{1.5cm}<{\centering}m{1.5cm}<{\centering}m{1.5cm}<{\centering}}
    \toprule
    \quad \quad Method & Macro F1 & Accuracy & AUC & spAUC & F1$_\text{real}$ & F1$_\text{fake}$ \\
    \hline
    \multicolumn{7}{c}{\textbf{Dataset: \textit{Weibo}}} \\
    BiGRU \cite{cho2014on} & 74.53{\color{grayv} $\pm$0.36} & 83.61{\color{grayv} $\pm$0.96} & 86.86{\color{grayv} $\pm$0.45} & 69.99{\color{grayv} $\pm$0.95} & 89.47{\color{grayv} $\pm$0.74} & 60.06{\color{grayv} $\pm$0.70} \\
    EANN \cite{wang2018eann} & 75.61{\color{grayv} $\pm$0.37} & 83.95{\color{grayv} $\pm$0.97} & 87.18{\color{grayv} $\pm$0.46} & 70.13{\color{grayv} $\pm$0.95} & 89.89{\color{grayv} $\pm$0.74} & 60.87{\color{grayv} $\pm$0.71} \\
    MDFEND \cite{nan2021mdfend} & 75.44{\color{grayv} $\pm$0.84} & 83.99{\color{grayv} $\pm$0.98} & 86.95{\color{grayv} $\pm$0.87} & 69.90{\color{grayv} $\pm$0.71} & 89.92{\color{grayv} $\pm$0.73} & 60.94{\color{grayv} $\pm$1.11} \\
    BERT-EMO \cite{zhang2021mining}& 76.11{\color{grayv} $\pm$0.95} & 84.32{\color{grayv} $\pm$0.66} & 87.61{\color{grayv} $\pm$1.12} & 70.97{\color{grayv} $\pm$0.72} & 90.19{\color{grayv} $\pm$0.44} & 61.23{\color{grayv} $\pm$1.54} \\
    CED \cite{wu2023category} & 75.96{\color{grayv} $\pm$0.32} & 84.38{\color{grayv} $\pm$0.63} & 87.40{\color{grayv} $\pm$0.91} & 70.39{\color{grayv} $\pm$0.55} & 90.18{\color{grayv} $\pm$0.54} & 61.74{\color{grayv} $\pm$0.85} \\
    DM-INTER \cite{wang2024why} & 75.83{\color{grayv} $\pm$0.73} & 84.07{\color{grayv} $\pm$0.46} & 87.12{\color{grayv} $\pm$0.56} & 70.11{\color{grayv} $\pm$0.73} & 89.87{\color{grayv} $\pm$0.46} & 61.58{\color{grayv} $\pm$0.83} \\
    
    \hline
    \textbf{BERT} \cite{devlin2019bert} & 75.53{\color{grayv} $\pm$0.32} & 84.28{\color{grayv} $\pm$0.33} & 87.19{\color{grayv} $\pm$0.46} & 70.69{\color{grayv} $\pm$0.47} & 90.16{\color{grayv} $\pm$0.28} & 60.91{\color{grayv} $\pm$0.74} \\
    \quad + ENDEF \cite{zhu2022generalizing} & 75.92{\color{grayv} $\pm$0.52} & 84.73{\color{grayv} $\pm$0.35} & 87.31{\color{grayv} $\pm$0.58} & 70.25{\color{grayv} $\pm$0.40} & 90.48{\color{grayv} $\pm$0.29} & 61.37{\color{grayv} $\pm$1.11} \\
    \quad + DER \cite{li2021prefix} & 75.09{\color{grayv} $\pm$0.32} & 84.20{\color{grayv} $\pm$0.33} & 86.78{\color{grayv} $\pm$0.46} & \underline{71.12}{\color{grayv} $\pm$1.16} & 90.16{\color{grayv} $\pm$0.28} & 60.02{\color{grayv} $\pm$0.74} \\
    \rowcolor{lightgrayv} \quad + \baby-\textsc{lstm} & 76.19{\color{grayv} $\pm$0.59}* & 84.77{\color{grayv} $\pm$0.49} & 87.94{\color{grayv} $\pm$0.84}* & 70.24{\color{grayv} $\pm$0.39} & 90.48{\color{grayv} $\pm$0.38} & 61.90{\color{grayv} $\pm$1.01}* \\
    
    \rowcolor{lightgrayv} \quad + \baby-\textsc{ode} & \underline{76.59}{\color{grayv} $\pm$0.25}* & \underline{85.37}{\color{grayv} $\pm$0.32}* & \underline{87.99}{\color{grayv} $\pm$0.73}* & 71.06{\color{grayv} $\pm$0.98} & \underline{90.95}{\color{grayv} $\pm$0.19}* & \underline{62.13}{\color{grayv} $\pm$0.50}* \\

    \rowcolor{lightgrayv} \quad + \baby-\textsc{pt} & \textbf{76.82}{\color{grayv} $\pm$0.79}* & \textbf{85.61}{\color{grayv} $\pm$0.63}* & \textbf{88.84}{\color{grayv} $\pm$0.80}* & \textbf{72.09}{\color{grayv} $\pm$0.63}* & \textbf{91.09}{\color{grayv} $\pm$0.39}* & \textbf{62.55}{\color{grayv} $\pm$0.78}* \\
    \hline
    \specialrule{0em}{0.5pt}{0.5pt}
    \hline
    
    \multicolumn{7}{c}{\textbf{Dataset: \textit{GossipCop}}} \\
    BiGRU \cite{cho2014on} & 77.24{\color{grayv} $\pm$0.57} & 83.26{\color{grayv} $\pm$0.65} & 85.57{\color{grayv} $\pm$0.41} & 73.24{\color{grayv} $\pm$0.43} & 88.94{\color{grayv} $\pm$0.56} & 65.56{\color{grayv} $\pm$0.99} \\
    EANN \cite{wang2018eann} & 78.52{\color{grayv} $\pm$0.69} & 84.66{\color{grayv} $\pm$0.43} & 87.36{\color{grayv} $\pm$0.51} & 75.47{\color{grayv} $\pm$0.51} & 90.00{\color{grayv} $\pm$0.30} & 67.05{\color{grayv} $\pm$1.20} \\
    MDFEND \cite{nan2021mdfend} & 78.62{\color{grayv} $\pm$0.49} & 84.59{\color{grayv} $\pm$0.38} & 87.65{\color{grayv} $\pm$0.42} & 75.90{\color{grayv} $\pm$0.33} & 89.92{\color{grayv} $\pm$0.31} & 67.32{\color{grayv} $\pm$0.91} \\
    BERT-EMO \cite{zhang2021mining} & 78.84{\color{grayv} $\pm$0.37} & 84.65{\color{grayv} $\pm$0.50} & 87.57{\color{grayv} $\pm$0.57} & 75.60{\color{grayv} $\pm$0.39} & 89.92{\color{grayv} $\pm$0.41} & 67.75{\color{grayv} $\pm$0.50} \\
    CED \cite{wu2023category} & 78.58{\color{grayv} $\pm$0.59} & 84.46{\color{grayv} $\pm$0.91} & 87.14{\color{grayv} $\pm$0.57} & 73.89{\color{grayv} $\pm$1.49} & 89.78{\color{grayv} $\pm$0.82} & 67.39{\color{grayv} $\pm$1.15} \\
    DM-INTER \cite{wang2024why} & 78.52{\color{grayv} $\pm$0.57} & 84.48{\color{grayv} $\pm$0.73} & 87.20{\color{grayv} $\pm$0.66} & 74.95{\color{grayv} $\pm$0.63} & 89.90{\color{grayv} $\pm$0.44} & 67.47{\color{grayv} $\pm$0.74} \\
    
    \hline
    \textbf{BERT} \cite{devlin2019bert} & 78.32{\color{grayv} $\pm$0.32} & 84.66{\color{grayv} $\pm$0.30} & 86.72{\color{grayv} $\pm$0.52} & 74.63{\color{grayv} $\pm$0.55} & 90.04{\color{grayv} $\pm$0.26} & 66.61{\color{grayv} $\pm$0.66} \\
    \quad + ENDEF \cite{zhu2022generalizing} & 79.01{\color{grayv} $\pm$0.73} & 84.89{\color{grayv} $\pm$0.58} & 86.98{\color{grayv} $\pm$0.53} & 75.02{\color{grayv} $\pm$0.66} & 90.11{\color{grayv} $\pm$0.43} & 67.91{\color{grayv} $\pm$1.22} \\
    \quad + DER \cite{li2021prefix} & 78.31{\color{grayv} $\pm$0.33} & 84.79{\color{grayv} $\pm$0.26} & 86.65{\color{grayv} $\pm$0.08} & 74.82{\color{grayv} $\pm$0.50} & 90.16{\color{grayv} $\pm$0.27} & 66.47{\color{grayv} $\pm$0.88} \\
    \rowcolor{lightgrayv} \quad + \baby-\textsc{lstm} & 79.18{\color{grayv} $\pm$0.30}* & 84.82{\color{grayv} $\pm$0.58} & 87.04{\color{grayv} $\pm$0.62} & 75.24{\color{grayv} $\pm$0.51}* & 90.20{\color{grayv} $\pm$0.35} & 68.32{\color{grayv} $\pm$0.75}* \\
    
    \rowcolor{lightgrayv} \quad + \baby-\textsc{ode} & \underline{79.67}{\color{grayv} $\pm$0.28}* & \underline{85.26}{\color{grayv} $\pm$0.30}* & \underline{87.57}{\color{grayv} $\pm$0.22}* & \underline{75.75}{\color{grayv} $\pm$0.35}* & \underline{90.31}{\color{grayv} $\pm$0.28} & \textbf{69.04}{\color{grayv} $\pm$0.51}* \\

    \rowcolor{lightgrayv} \quad + \baby-\textsc{pt} & \textbf{79.98}{\color{grayv} $\pm$0.49}* & \textbf{85.43}{\color{grayv} $\pm$0.37}* & \textbf{87.91}{\color{grayv} $\pm$0.46}* & \textbf{76.56}{\color{grayv} $\pm$0.69}* & \textbf{90.83}{\color{grayv} $\pm$0.47}* & \underline{68.95}{\color{grayv} $\pm$0.79}* \\
    \bottomrule
  \end{tabular} }
\end{table*}

\subsection{Main Results and Analysis}

We conduct a comparative analysis of our proposed three variants of \baby-\textsc{lstm}, \baby-\textsc{ode}, and \baby-\textsc{pt} against several baseline models. Their experimental results are reported in Table~\ref{result}. To ensure the credibility of the experimental results, the experiments for each model are repeated five times with five different random seeds \{1, 2, 3, 4, 5\}, and their average scores and standard deviations are presented finally. Following previous works \cite{zhu2022generalizing}, we utilize six widely used metrics to evaluate the performance of these methods. The employed metrics include macro F1, accuracy (Acc.), Area Under Curve (AUC), standardized sparse AUC (spAUC), as well as the F1 scores for real and misinformation (F1$_\text{real}$ and F1$_\text{fake}$).
The experimental results generally indicate that all three variants consistently surpass the baseline methods across all evaluation metrics. Our \baby method is designed to be plug-and-play, and compared to the backbone detector BERT, \baby can significantly enhance its performance, and the results on most metrics show statistically significant improvements over the baseline model. For instance, we observe improvements of 0.93 and 1.64 in the F1 scores for real and misinformation, respectively, using \baby-\textsc{ode} on the \textit{Weibo} dataset. Compared to the state-of-the-art method ENDEF, which also addresses the time-evolving nature of MD, \baby also demonstrates competitive results. For example, it achieves improvements of 0.90 and 0.88 in terms of Macro F1 and Acc. on the \textit{Weibo} dataset. 
This comparison highlights the effectiveness of temporal modeling approaches such as ENDEF and \baby when applied in scenarios where the model is trained on historical data and evaluated on future data.
Moreover, the introduction of our time series model, which incorporates social environmental features, provides further enhancements in predictive performance. By modeling the dynamic evolution of the social environment, this approach allows \baby to capture long-term trends and subtle contextual shifts, making it particularly well-suited for real-world MD tasks.

Turning to compare our three variants, the experimental results and their corresponding standard deviations clearly indicate the following ranking: \baby-\textsc{pt} $>$ \baby-\textsc{ode} $>$ \baby-\textsc{lstm}. This result highlights several important insights. First, the continuous-time dynamical models \baby-\textsc{ode} and  \baby-\textsc{pt} consistently outperform the discrete-time time series model \baby-\textsc{lstm}, demonstrating the advantage of modeling temporal dynamics in a continuous framework rather than relying on discrete time steps. Second, the pre-trained \baby-\textsc{pt} variant exhibits the strongest generalization and predictive capabilities. This suggests that incorporating pre-training enhances the model's ability to adapt to a wide range of data and contexts, leading to better performance and greater robustness across different evaluation metrics.

\subsection{Sensitivity Analysis}

In this section, we conduct a sensitivity analysis on two crucial hyper-parameters to facilitate parameter selection.

\subsubsection{Varying Time Intervals}

\begin{figure}[t]
  \centering
  \includegraphics[scale=0.33]{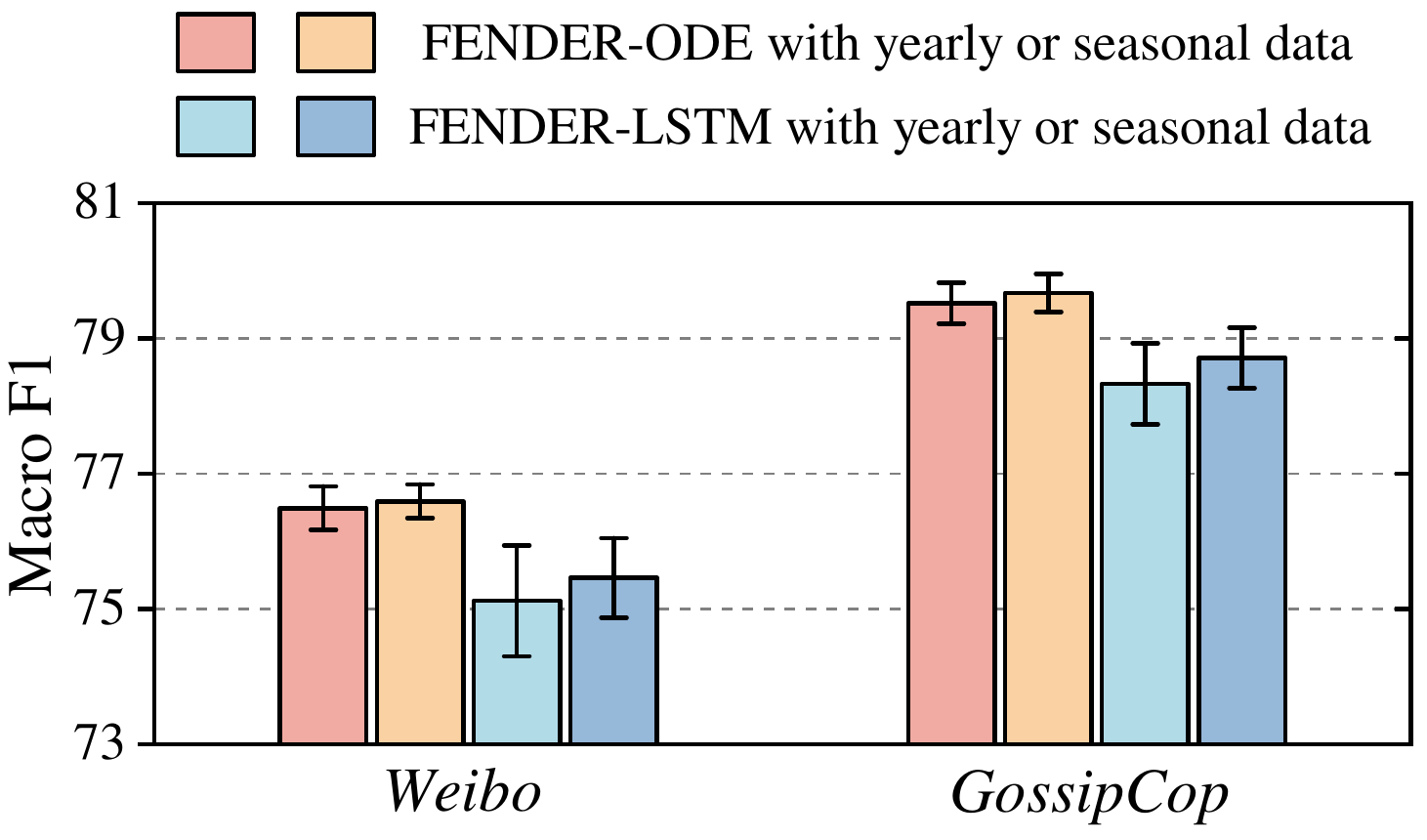}
  \caption{Sensitivity analysis of the time interval, \ie yearly and seasonal intervals.}
  \label{sensitivityT}
\end{figure}

During the training process of the time series model, we partition the complete dataset $\mathcal{D}$ into $T$ subsets $\{\mathcal{D}^\tau\}_{\tau=1}^{T}$ based on the time stamps. In this section, we empirically investigate the impact of different partitioning granularities on model performance. Specifically, we explore various approaches, such as partitioning the data into subsets representing one season (three months) or one full year.
To evaluate these partitioning strategies, we conduct experiments with two variants \baby-\textsc{lstm} and \baby-\textsc{ode}, which are representatives of discrete and continuous time models respectively, on two datasets \textit{GossipCop} and \textit{Weibo}. The experimental results are reported in Fig.~\ref{sensitivityT}. 

The experimental results reveal a clear and consistent trend: models trained with seasonal data (i.e., data segmented into finer, more frequent time intervals) consistently outperform those trained with coarser, yearly time periods. This observation underscores the potential benefits of utilizing fine-grained temporal data to train more accurate and effective time series models. It suggests that models can capture more nuanced temporal patterns and dynamics when trained on data with higher time resolution, leading to improved performance in MD.
Additionally, we observe that \baby-\textsc{lstm} is more sensitive to the length of the time interval than \baby-\textsc{ode}. This sensitivity indicates that the discrete nature of the LSTM model makes it more susceptible to variations in time granularity. In contrast, \baby-\textsc{ode}, which leverages continuous dynamics equations, demonstrates greater robustness to changes in input data. The continuous-time framework of \baby-\textsc{ode} appears to better capture long-term dependencies and subtle shifts in the temporal evolution of the data, making it less sensitive to the specific choice of time intervals.

\begin{figure}[t]
  \centering
  \includegraphics[scale=0.37]{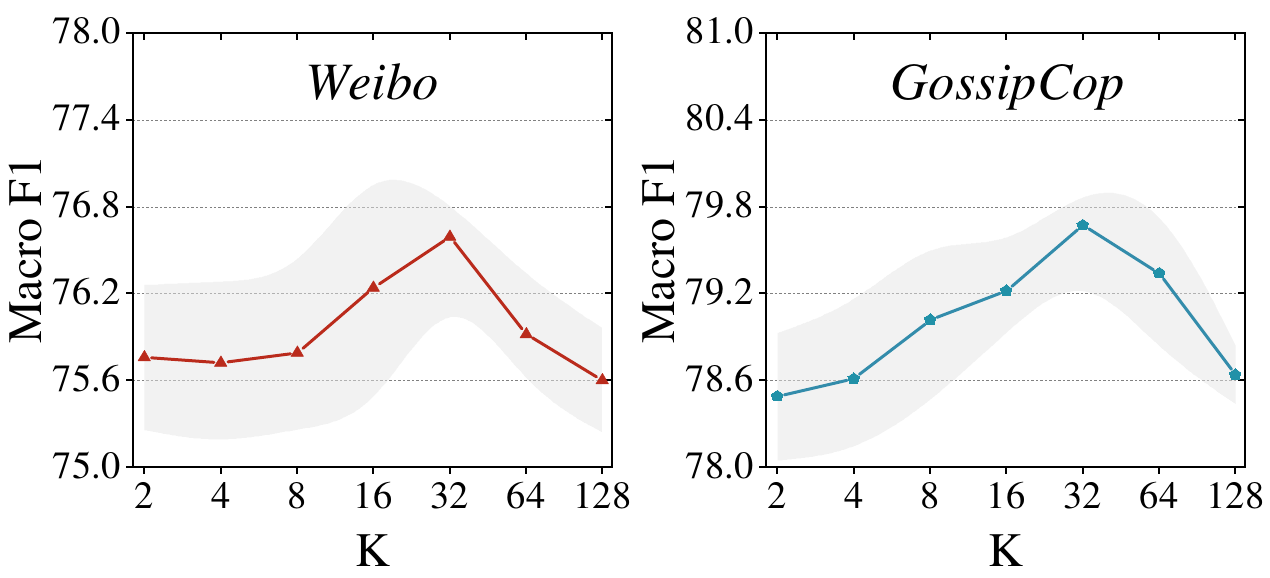}
  \caption{Sensitivity analysis of the sequence length $K$.}
  \label{sensitivityK}
\end{figure}

\definecolor{bluev}{HTML}{3865B6}
\begin{table*}[t]
\centering
\renewcommand\arraystretch{1.0}
  \caption{Long-term prediction results of \baby. {\footnotesize $\uparrow$} and {\footnotesize $\downarrow$} denotes the gap between the results compared to the 0\% drop rate.}
  \label{missing}
  \small
  \setlength{\tabcolsep}{5pt}{
  \begin{tabular}{m{0.7cm}<{\centering}m{2.0cm}<{\centering}m{1.4cm}<{\centering}m{1.4cm}<{\centering}m{1.4cm}<{\centering}m{1.4cm}<{\centering}m{1.4cm}<{\centering}m{1.4cm}<{\centering}m{1.0cm}<{\centering}}
    \toprule
    Rate & Model & Macro F1 & Acc. & AUC & spAUC & F1$_\text{real}$ & F1$_\text{fake}$ & Avg.~$\Delta$ \\
    \hline
    \multirow{2}{*}{\textbf{0 \%}} & \baby-\textsc{pt} & 79.98 & 85.43 & 87.91 & 76.56 & 90.83 & 68.95 & - \\
    & \baby-\textsc{ode} & 79.67 & 85.26 & 87.57 & 75.75 & 90.31 & 69.04 & - \\
    & \baby-\textsc{lstm} & 79.18 & 84.82 & 87.04 & 75.24 & 90.20 & 68.32 & - \\
    
    \hline
    \multirow{2}{*}{10 \%} & \baby-\textsc{pt} & 79.63{\color{grayv} $\downarrow$0.35} & 85.37{\color{grayv} $\downarrow$0.06} & 87.79{\color{grayv} $\downarrow$0.12} & 75.72{\color{bluev} $\downarrow$\textbf{0.84}} & 90.63{\color{grayv} $\downarrow$0.20} & 68.63{\color{grayv} $\downarrow$0.32} & \textbf{-0.315} \\
    & \baby-\textsc{ode} & 79.67{\color{grayv} $\downarrow$0.00} & 84.35{\color{bluev} $\downarrow$\textbf{0.91}} & 87.22{\color{grayv} $\downarrow$0.35} & 75.85{\color{grayv} $\uparrow$0.10} & 89.42{\color{bluev} $\downarrow$\textbf{0.89}} & 69.93{\color{grayv} $\uparrow$0.89} & \textbf{-0.193} \\
    & \baby-\textsc{lstm} & 78.45{\color{bluev} $\downarrow$\textbf{0.72}} & 84.01{\color{grayv} $\downarrow$0.81} & 86.45{\color{bluev} $\downarrow$\textbf{0.59}} & 74.57{\color{grayv} $\downarrow$0.67} & 89.40{\color{grayv} $\downarrow$0.80} & 67.45{\color{bluev} $\downarrow$\textbf{0.87}} & \textbf{-0.743} \\
    
    \hline
    \multirow{2}{*}{30 \%} & \baby-\textsc{pt} & 79.16{\color{grayv} $\downarrow$0.82} & 84.59{\color{grayv} $\downarrow$0.84} & 87.63{\color{grayv} $\downarrow$0.28} & 75.62{\color{bluev} $\downarrow$\textbf{0.94}} & 89.80{\color{grayv} $\downarrow$1.03} & 68.51{\color{grayv} $\downarrow$0.44} & \textbf{-0.725} \\
    & \baby-\textsc{ode} & 79.08{\color{grayv} $\downarrow$0.59} & 83.73{\color{grayv} $\downarrow$1.53} & 86.54{\color{bluev} $\downarrow$\textbf{1.03}} & 75.87{\color{grayv} $\uparrow$0.12} & 88.94{\color{grayv} $\downarrow$1.37} & 69.22{\color{grayv} $\uparrow$0.18} & \textbf{-0.703} \\
    & \baby-\textsc{lstm} & 78.07{\color{bluev} $\downarrow$\textbf{1.11}} & 82.51{\color{bluev} $\downarrow$\textbf{2.31}} & 86.37{\color{grayv} $\downarrow$0.67} & 76.25{\color{grayv} $\uparrow$1.01} & 87.93{\color{bluev} $\downarrow$\textbf{2.27}} & 67.80{\color{bluev} $\downarrow$\textbf{0.52}} & \textbf{-0.978} \\
    
    \hline
    \multirow{2}{*}{50 \%} & \baby-\textsc{pt} & 78.65{\color{grayv} $\downarrow$1.33} & 83.95{\color{grayv} $\downarrow$1.48} & 86.39{\color{grayv} $\downarrow$1.52} & 75.65{\color{bluev} $\downarrow$\textbf{0.91}} & 89.29{\color{grayv} $\downarrow$1.54} & 68.02{\color{grayv} $\downarrow$0.93} & \textbf{-1.285} \\
    & \baby-\textsc{ode} & 78.51{\color{grayv} $\downarrow$1.16} & 83.95{\color{grayv} $\downarrow$1.31} & 86.04{\color{bluev} $\downarrow$\textbf{1.53}} & 76.00{\color{grayv} $\uparrow$0.25} & 88.95{\color{grayv} $\downarrow$1.36} & 68.07{\color{grayv} $\downarrow$0.97} & \textbf{-1.013} \\
    & \baby-\textsc{lstm} & 77.27{\color{bluev} $\downarrow$\textbf{1.91}} & 81.61{\color{bluev} $\downarrow$\textbf{3.21}} & 85.55{\color{grayv} $\downarrow$1.49} & 75.68{\color{grayv} $\uparrow$0.44} & 87.20{\color{bluev} $\downarrow$\textbf{3.00}} & 67.11{\color{bluev} $\downarrow$\textbf{1.21}} & \textbf{-1.730} \\
    \bottomrule
  \end{tabular} }
\end{table*}

\subsubsection{Varying Sequence Length $L$}

The length $L$ of our designed DER is a crucial hyper-parameter in our model. In this section, we conduct a sensitivity analysis to examine the impact of this parameter on model performance. Specifically, we experiment with \baby-\textsc{ode} on two datasets, selecting values for $L$ from the set \{2, 4, 8, 16, 32, 64, 128\}.  The corresponding macro F1 scores for each configuration are presented in Fig.~\ref{sensitivityK}.
Interestingly, the results reveal a clear pattern: the model consistently achieves its best performance when $L = 32$. Beyond this point, whether $L$ increases or decreases, performance on both datasets shows a noticeable decline.
We offer two main explanations for this observation. First, smaller sequence lengths (i.e., lower values of $L$) are insufficient for effectively capturing the complexity and evolution of the social environment over time. The social dynamics, which are central to our model, require a minimum sequence length to adequately reflect these shifts. Second, excessively large sequence lengths place a significant burden on the time series modeling process. When $L$ becomes too large, the model struggles with increased computational complexity and potentially suffers from overfitting, as it may start to capture irrelevant noise or lose focus on key temporal patterns.

\subsection{Long-term Prediction}

Typically, time series models are evaluated through long-term prediction analysis, which assesses the model's ability to make accurate predictions over an extended time span, with a significant gap between the training and test datasets. In this section, we conduct experiments on the \textit{GossipCop} dataset to investigate the long-term prediction capabilities of the three variants of our model.
To carry out this analysis, we first sort the training dataset in ascending order based on time, ensuring that the temporal order is preserved. Then, we systematically discard a portion of the data from the tail end of the dataset, simulating the scenario where future data is withheld for testing. The proportion of data to be removed, or the drop rate, is selected from the set $\{10\%, 30\%, 50\%\}$. The results are presented in Table~\ref{missing}, and we report the average decline (Avg. $\Delta$) to illustrate the average decreases in performance across all metrics.

As the drop rate increases, the performance of these variants consistently declines, highlighting the challenges posed by limited training data. At a drop rate of 10\%, both variants \baby-\textsc{lstm} and \baby-\textsc{ode} exhibit similar levels of performance, with no significant difference observed between them. However, as the drop rate rises to 30\% or 50\%, the performance of \baby-\textsc{lstm} experiences a notable and more rapid decline compared to \baby-\textsc{ode} and \baby-\textsc{pt}.
These findings indicate that \baby-\textsc{ode} and \baby-\textsc{pt}, with their continuous dynamics equations, demonstrate superior capabilities in handling long-term predictions and exhibit greater resilience when confronted with data scarcity. The continuous nature of these models allows them to better capture and extrapolate from the available data, even when significant portions are missing. In contrast, \baby-\textsc{lstm} appears to be more susceptible to data gaps, suggesting that the discrete nature of the LSTM model constrains its ability to generalize over extended time periods. The discrete steps in the LSTM's processing mechanism may lead to a loss of information continuity, which is crucial for maintaining accuracy in scenarios with high drop rates.

\begin{figure*}[h]
  \centering
  \includegraphics[scale=0.54]{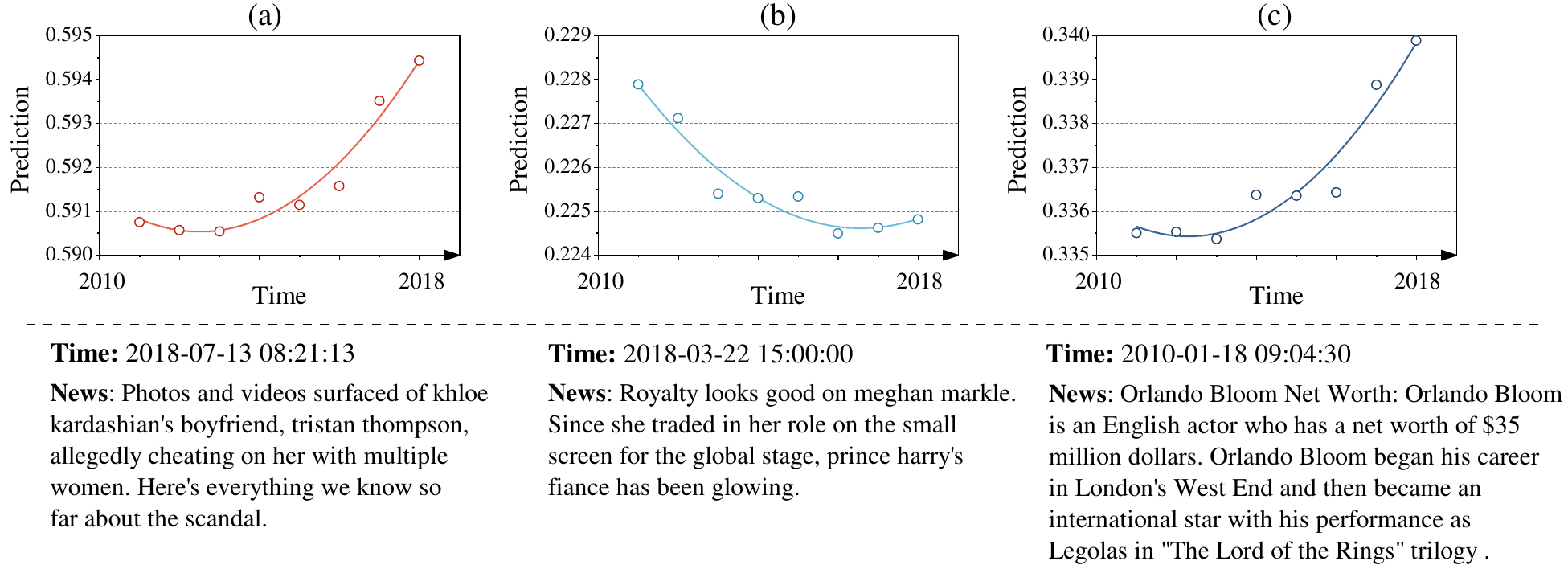}
  \caption{We illustrate three representative examples for the case study of the \baby-\textsc{ode} variant.}
  \label{casestudy}
\end{figure*}

\begin{table}[t]
\centering
\renewcommand\arraystretch{1.10}
  \caption{Time costs of MD models on two datasets.}
  \label{time}
  \setlength{\tabcolsep}{5pt}{
  \begin{tabular}{m{1.3cm}<{\centering}m{0.6cm}<{\centering}m{1.8cm}<{\centering}m{1.6cm}<{\centering}m{1.5cm}<{\centering}}
    \toprule
    Time & BERT & \baby-\textsc{lstm} & \baby-\textsc{ode} & \baby-\textsc{pt} \\
    \hline
    \textit{Weibo} & 11.7 & 13.3 (1.13$\times$) & 14.5 (1.23$\times$) & 18.0 (1.53$\times$) \\
    \textit{GossipCop} & 12.2 & 14.6 (1.19$\times$) & 15.6 (1.27$\times$) & 17.9 (1.46$\times$) \\
    \bottomrule
  \end{tabular} }
\end{table}

\subsection{Time Cost Analysis}

Compared to the process of training a standard detector, our proposed \baby framework incorporates an additional dynamic environment learning phase, which can potentially increase computational costs. To thoroughly evaluate this aspect, we conduct a time cost analysis in this section. Specifically, we record the runtime performance of BERT and our three variants across two datasets. The average runtimes for five experiments, each using five different random seeds, are presented in Table~\ref{time}.
The experimental results reveal a clear trend: as the scale of the time series model increases, the runtime of the model also progressively increases. Notably, the most complex variant \baby-\textsc{pt} of our framework exhibits a runtime that is up to 1.53 times longer than that of the baseline BERT model, which is acceptable. This significant increase in computational time, however, is accompanied by a corresponding improvement in model accuracy. Specifically, the accuracy of \baby-\textsc{pt} is enhanced by a factor of 1.33 compared to the baseline. This suggests that the additional time cost is justified by the substantial performance gains, making the trade-off between computational resources and model performance a reasonable one.

\subsection{Case Study}

During the dynamic environment learning process in our model, we focus on learning a temporal-evolving DER while keeping all other parameters fixed. This approach enables the \baby framework to generate a distinct MD model for each time period, allowing it to adapt to the evolving nature of the social environment. To explore this inherent dynamic, we present three representative cases, shown in Fig.~\ref{casestudy}, that illustrate how the model’s predictions evolve over time.

In this experiment, we provide veracity predictions for each time period and track how these predictions change as time progresses. The cases are drawn from the \textit{GossipCop} dataset, which spans from 2000 to 2017. Specifically, cases (a) and (b) are from 2018, with their veracity labels being fake and real, respectively. Over time, we observe that the predictions for these cases converge more closely with the ground-truth labels, indicating that the model accurately captures the temporal dynamics and shifting social context. This behavior highlights the ability of our approach to adapt to the evolving nature of news dissemination and public opinion.
In contrast, case (c) is from 2010 and has a veracity label of real. Interestingly, the model's predictions for this case diverge significantly from the ground-truth label as time progresses. This anomaly underscores the challenges associated with long-term prediction and suggests that the social context surrounding a news piece may have shifted considerably over time, leading to a reevaluation of its veracity.

\section{Conclusion and Future Works} \label{sec5}

In this paper, we focus on capturing the dynamically evolving social context, which plays a crucial role in affecting the veracities of news articles. To achieve this goal, we propose a novel temporal MD framework named \baby. Our approach begins by employing a prompt tuning technique as the social contextual representation and pre-training the misinformation detector using the complete dataset. We then efficiently fine-tune the representation with data from each time period while keeping the other parameters frozen. Concurrently, we learn a time series model, such as an LSTM model or a continuous-time dynamics equation, using the fine-tuned representations as ground-truth labels. The ultimate aim is to predict a new representation for the future period, facilitating the detection process.
Our experimental results demonstrate that \baby outperforms state-of-the-art MD baselines across two datasets. Additionally, we conduct further experiments to investigate the continuous nature of the dynamics equation.

\vspace{2pt} \noindent
\textbf{Limitations.}
To be honest, our \baby framework does have certain limitations. One of the notable issues is the imbalanced numbers of samples in the training dataset for each time period. For instance, in the \textit{GossipCop} dataset, we observe only 12 samples for summer in 2010, whereas there are 1,700 samples for winter in 2017. This imbalance in the data unavoidably affects the performance of the fine-tuned social contextual representations.

\section*{Acknowledgement}
We acknowledge support for this project from the National Key R\&D Program of China (No.2021ZD0112501, No.2021ZD0112502), the National Natural Science Foundation of China (No.62276113), and China Postdoctoral Science Foundation (No.2022M721321).



\bibliographystyle{ACM-Reference-Format}
\balance
\bibliography{reference}


\end{document}